\documentclass[conference]{IEEEtran}
\IEEEoverridecommandlockouts
\usepackage[nocompress]{cite}
\usepackage{amsmath,amssymb,amsfonts}
\usepackage[numbers,sort&compress]{natbib}
\usepackage{algorithmic}
\usepackage{graphicx}
\usepackage{textcomp}
\usepackage{xcolor}
\usepackage{amsmath}

\def\BibTeX{{\rm B\kern-.05em{\sc i\kern-.025em b}\kern-.08em
    T\kern-.1667em\lower.7ex\hbox{E}\kern-.125emX}}

\usepackage{hyperref}
\hypersetup{
colorlinks   = true,
citecolor    = blue,
urlcolor    =  blue
}
\usepackage{url}
\usepackage{multirow}
\usepackage{hhline}

\newcommand{\footURL}[1]{\footnote{\url{#1}}}

\makeatletter
\newcommand\footnoteref[1]{\protected@xdef\@thefnmark{\ref{#1}}\@footnotemark}
\makeatother

\newcommand{\dnd}{D\&D}

\begin{document}

\title{Fine Tuning Named Entity Extraction Models for the Fantasy Domain}

\author{\IEEEauthorblockN{Aravinth Sivaganeshan, Nisansa de Silva}
\IEEEauthorblockA{\textit{Department of Computer Science and Engineering,} \\
\textit{University of Moratuwa}\\ Moratuwa, Sri Lanka\\
\{sivaganeshan.22,nisansadds\}@cse.mrt.ac.lk}
}

\IEEEpubid{\makebox[\columnwidth]{979-8-3503-4521-6/23/\$31.00\copyright2023 IEEE\hfill} 
\hspace{\columnsep}\makebox[\columnwidth]{ }}



\maketitle

\IEEEpubidadjcol

\begin{abstract}

Named Entity Recognition (NER) is a sequence classification Natural Language Processing task where entities are identified in the text and classified into predefined categories. It acts as a foundation for most information extraction systems. Dungeons and Dragons (\dnd) is an open-ended tabletop fantasy game with its own diverse lore. DnD entities are domain-specific and are thus unrecognizable by even the state-of-the-art off-the-shelf NER systems as the NER systems are trained on general data for pre-defined categories such as: person (\verb|PERS|), location (\verb|LOC|), organization (\verb|ORG|), and miscellaneous (\verb|MISC|). For meaningful extraction of information from fantasy text,  the entities need to be classified into domain-specific entity categories as well as the models be fine-tuned on a domain-relevant corpus. This work uses available lore of monsters in the \dnd domain to fine-tune \textit{Trankit}, which is a prolific NER framework that uses a pre-trained model for NER. Upon this training, the system acquires the ability to extract monster names from relevant domain documents under a novel NER tag. This work compares the accuracy of the monster name identification against; the zero-shot \textit{Trankit} model and two \textit{FLAIR} models.  The fine-tuned \textit{Trankit} model achieves an 87.86\%  F1 score surpassing all the other considered models.
\end{abstract}

\renewcommand\IEEEkeywordsname{Keywords}


\section{Introduction}
Named Entity Recognition is a foundation-level NLP task that aims to recognize mentions of rigid designators belonging to syntactic types such as \textit{person}, \textit{location}, \textit{organization}~\cite{nadeau2007survey}. This subsequently serves as the foundation upon which other tasks such as coreference resolution, relation extraction and event extraction are built. Further, some higher-level tasks such as text summarization, text understanding, information retrieval, question answering, and machine translation may use tagged NER as one of the inputs. From its inception, NER task has evolved with the technologies of the time; broadly including rule/knowledge-based, unsupervised, feature-engineered and supervised, feature-inferring and deep learning~\cite{yadav2019survey,li2020survey}. The state-of-the-art methods use deep learning methods and are based on models which are trained using large amounts of data; such as BERT~\cite{devlin2018bert} which is a transformer-based pre-trained model which is then fine-tuned on the specific NLP task. However, such NER systems are fine-tuned on the general domain with the objective of high generalizability. This in turn sacrifices specificity on domain-specific applications~\cite{sugathadasa2017synergistic,wijeratne2019natural} as some general categories may not be applicable as well as some specific categories may not exist.

Dungeons and Dragons (\dnd) is a tabletop, open-ended, fantasy role-playing game that has been commercially available since 1974. \dnd{} has a pre-defined set of rules covering the gameplay~\cite{peiris2022synthesis,squire2007open,peiris2023shade}. Also for Dungeons and Dragons, there are several settings and each setting has lore (historical and current status of the setting). Over the decades, several updates and improvements have been made to the game and thus, currently, it is in its fifth version~\cite{crawford2014player}. In a \dnd{} game, combat encounters are one of the most important components as they are the means by which the player character attains progress~\cite{master2014dungeon}. The opponents that are pitted against the players in these encounters are referred to as \textit{monsters} regardless of the general domain dictionary definition of the term. For example, a \textit{Bandit} the players encounter is tagged as a \textit{monster} just as it would tag a \textit{dragon} or \textit{unicorn}. This domain-specific definition poses a unique challenge to NLP tasks. Further, when generating encounters in \dnd, it is preferred that the encounters are aligned with the lore to conserve immersion and verisimilitude~\cite{stern2002touch}. Thus, it is vital to extract information from the lore pertaining to monsters, their accompanying monsters, their terrain and their habitat before generating encounters in \dnd. NER for this fantasy domain is useful as it is the foundation for the information extraction and also the identified monster names can be used for the generation of encounters in \dnd.

The models that are currently being used for the general domain cannot be directly used~\cite{tang2017nite} for the fantasy domain, as identifying the generic categories from fantasy domain text might not be useful, as well as there is no extant provision to extract task-specific categories.
Thus, these pre-trained models should be fine-tuned with fantasy domain-specific data for the custom categories. In this study, the monster lore data for the \dnd{} monsters are extracted and monster entities are tagged using text lookup. Then the dataset that is tagged in \verb|BIO| format is obtained and the custom NER training of \textit{Trankit}~\cite{van2021trankit} and \textit{FLAIR}~\cite{akbik2019flair} are utilized for fine-tuning. The trained models are then evaluated using precision, recall and F1. Then, predictions are done using the model on the complete lore data and the relationships between the monsters are mapped by an association map(list) based on the mention of a monster in another monster's lore. Finally, a directed graph is created for visualizing the association.

The remainder of this paper is organized as follows: Section~\ref{sec:related} discusses the existing work and the reason for selecting our approach. Section~\ref{sec:methodology} explains the 
methodology used in pre-processing the data and fine-tuning the model. Section~\ref{sec:results}
 discusses the results. And finally, Section~\ref{sec:conc} concludes the paper. 


\section{Related Works}
\label{sec:related}

Named entity recognition can be considered as one of the main tasks related to Information Extraction and methods for named entity recognition methods have evolved through several phases from rule-based and unsupervised learning approaches to deep learning-based approaches. Also, state-of-the-art methods for NER have moved recently from word, character-based representations and manually engineered features~\cite{passos2014lexicon,chiu2016named} to approaches that use deep learning-based contextualized representations~\cite{akbik2018contextual,peters-etal-2018-deep,devlin2018bert,baevski2019cloze}. 

Initial approaches such as rule-based and manually engineered approaches cannot be used for different tasks or different domains~\cite{yadav2019survey}. This is mainly due to the reason that rules need to be formulated or features need to be engineered according to the domain and the NER categories. Therefore, the models used with manually engineered features cannot be used with another domain. Additionally, the Approaches similar to the one proposed by~\citet{passos2014lexicon} use lexicon-infused embeddings and therefore, they rely on lexical features. Therefore it is not possible to apply or tune the above approach according to different NER categories or different domains.

In recent years, several deep-learning approaches have influenced NLP and in particular named entity recognition~\cite{luoma2020exploring}. Approaches such as proposed by~\citet{chiu2016named} use LSTM-based architecture with character/word representations. Also, there are methods~\cite{peters-etal-2018-deep,devlin2018bert,baevski2019cloze} which use pre-trained transformer-based models. Methods using pre-trained transformer models provide several advantages other than providing state-of-the-art performance when compared to other deep learning architectures~\cite{li2020survey}. Pre-trained models like BERT~\cite{devlin2018bert}, provide many other advantages. These models can be fine-tuned using task-specific labelled datasets. Also, these models can be used to model longer distance relationships within the text and also can provide contextualized representations of text~\cite{luoma2020exploring}. As they are pre-trained on large corpora, these models capture the language representations well enough that they can provide better performance when fine-tuned in any domain for the pre-trained languages.

Further, there are NLP toolkits that are available for different NLP tasks and make use of the mentioned deep learning models. spaCy\footURL{https://spacy.io/}, UDify~\cite{kondratyuk201975}, FLAIR~\cite{akbik2019flair}, CoreNLP~\cite{manning2014stanford}, UDPipe~\cite{straka2016udpipe}, Trankit~\cite{van2021trankit} and Stanza~\cite{qi2020stanza} are some of the toolkits. Out of the above toolkits \textit{Stanza}, \textit{Flair} and \textit{Trankit} provide support for customized NER. Each of these toolkits make use of state-of-the-art models which are competitive with one another in terms of performance on CoNLL03 dataset for English. \textit{FLAIR} has a \verb|ModelTrainer| that can be used to train supported models and also fine-tune pre-trained models. \textit{FLAIR} has an implementation of Contextual String embeddings~\cite{akbik2018contextual} as well as FLERT~\cite{schweter2020flert}, which is sequence tagger which has inputs of document-level contextual representations from pre-trained \textit{BERT}. It also provides model training and hyperparameter selection routines to facilitate training and testing workflows. One of the disadvantages of using Flair as an NLP toolkit is that it cannot process raw text and depends on external tokenizers. Stanza provides an NER system-based on contextualized string representation-based sequence tagger which uses the method proposed by~\citet{akbik2018contextual} and this model uses BiLSTM as a language model with BiLSTM-CRF as sequence tagger. Stanza also provides a tokenizer so, that it can be used completely for the training and testing workflow. Trankit \cite{van2021trankit} is another toolkit that has a complete pipeline from raw text including a tokenizer. \textit{Trankit} uses XLM-Roberta encoder~\cite{conneau2019unsupervised}, which is a multilingual transformer-based encoder which is shared across all NLP tasks. According to~\citet{van2021trankit}, Trankit has competitive performance with \textit{FLAIR} as well as \textit{Stanza} in NER in English. All of these 3 frameworks provide competitive performance in terms of F1 for NER for OntoNotes~\cite{pradhan2007ontonotes} and CoNLL03~\cite{sang2003introduction} datasets. This paper considers \textit{FLAIR} and \textit{Trankit} for comparison as the state-of-the-art model available in \textit{Stanza} is the same as the contextualized string representation-based sequence tagger in \textit{FLAIR}. Recent attempts to use LLMs for NER is discussed by~\citet{weerasundara2023comparative} but it is out of the scope of this study. 

\section{Methodology}
\label{sec:methodology}

Identification of named entities belonging to the category of monsters from the fantasy domain data using available models in \textit{Trankit} and \textit{FLAIR} is taken as the task to experiment with the fine-tuning of entity extraction models for the fantasy domain.

\subsection{Data}

The data is expected to be having several mentions of a variety of monsters and it was decided that combining monster lore data of several monsters would be a suitable approach to get this kind of data. There were several sources of this data and the data needs to be selected. The monster list from the DnD 5e System Reference Document was taken as the standard list of monsters for which the description data is taken for the task. The Open Game Licence (OGL) monster list was obtained from a free online source\footURL{https://5thsrd.org/gamemaster\_rules/monster \_indexes/monsters\_by\_cr}. This was the initial list of 317 monsters. For this list of monsters, the publicly available lore was obtained from DnDbeyond\footURL{https://www.dndbeyond.com/} for each monster where lore was available for 246 out of 317 monsters. The obtained data is converted to json format with the monster name as the key and monster lore texts as the value. 
\begin{figure}[htbp]
\centerline{\includegraphics[width=0.5\textwidth]{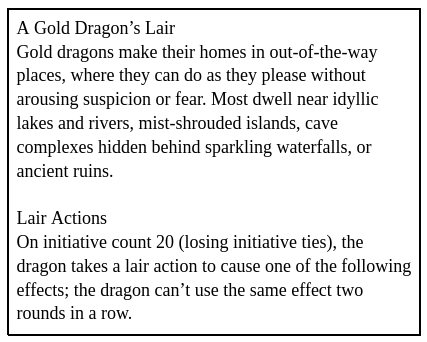}}
\caption{A part of the lore for ancient gold dragon}
\label{fig}
\end{figure}

Also, Forgotten Realms is considered the most popular setting in DnD. The Forgotten Realms Wiki data is extracted and presented in several formats in~\citet{peiris2022synthesis}. From this, the Wiki infobox data (FRW-I) was one of the data formats presented in~\citet{peiris2022synthesis}. \textit{FRW-I} data contains the Forgotten Realms Infobox data in json format. It was observed that the info boxes for the pages of monsters and spells had the key \verb|type5e|. Then by filtering the \verb|type5e| values for spells and getting the pages that had the \verb|type5e| values other than the filtered values, an additional list of monsters was made. The additional list contained 753 monsters. Then both the monster lists were merged and duplicates were removed to get the merged list of monsters that contained 853 monsters.

\subsection{Monster Tagging and converting to BIO format}

Obtained data contained monster names and the lore data in json format, the initial monster list and the merged monster list. There were 2 setups to try to create a dataset containing the tagged monsters. 

\begin{itemize}
    \item \textbf{Setup 1:} The initial monster list was to be looked up in the combined lore data obtained by combining all the monster lore from the lore data in the json format.
    \item \textbf{Setup 2:} The merged monster list was to be looked up in the combined lore data obtained by combining all the monster lore from the lore data in the json format.
\end{itemize}

Text lookup was done for monsters in the lore data as an initial step. When using the text look-up, there were issues with the false identification. For example, \textit{imp} is a monster and when using text lookup, \textit{imp} in words like \textit{impressive} will be highlighted as a mention of that monster. This will result in \textit{impressive} being tagged as a monster entity in the dataset which is incorrect. Also, \textit{ape} is a monster and \textit{ape} in words like \textit{shape} will be highlighted as the mention of the monster. There can be scenarios where the name of the monster can be a part of some other words, which are not actual mentions of that monster. The following steps were taken to solve this issue.

\begin{itemize}
    \item An association map was created using the text lookup for the monsters and the lore in which they appear. For example, if \textit{Ettin} appears in the lore of \textit{Green Dragon Wyrmling} and \textit{Goblin}. There would be an entry \verb|'Ettin':['Green Dragon Wyrmling',| \verb|'Goblin']|. 
    \item From the association map, it could be identified that the monsters that appeared in a larger number of lore were mostly related to the reason that they were a part of other words and not the actual mention of those monsters. 
    \item Therefore, from the observations, a threshold of 30 was taken and the list of monsters that had appeared in more than 30 lore were listed and made as the ignore list.
    
\end{itemize}

The above step of getting the ignore list was done separately for setup 1 and setup 2 where the text look-up was done separately for the initial list of monsters and the merged list of monsters respectively. Also, the initial list and the merged list of monsters were updated by removing the monsters from the ignore list.

As the first step to tokenize the data to \verb|BIO| format needed by the models, both the monster lists, that is the initial list and the merged list are sorted according to the length of monster names and sorted lists\footURL{https://github.com/aravinthnlp/DnDMonsters} were obtained for the initial list and the merged list. This is to search for the longer monster names first so that they are first tagged. This enables leaving out short monster names if a span of text contains more than one monster name. For example, if a sentence has \textit{steam mephit} where both \textit{steam mephit} and \textit{mephit} are monsters, both cannot be tagged. For leaving out the shortest out of the 2, the monster lists are sorted in descending order of the monster lengths. 

The combined lore data is sentence tokenized and then sentences are word tokenized using the \textit{Trankit} tokenizer and separated into sentences as the next step. For each sentence, a lookup was done for the found monsters in the order of the monster list and for every lookup, the index of the starting and ending of the monster is stored. When looking up for the next monster, the next monster is ignored if it is also found overlapping with the span of an already found monster. After the indexes of the found monsters are stored in the sentences, the word tokens in the sentences are tagged with \verb|B-MONS|, \verb|I-MONS| or \verb|O| according to whether the span of the token is at the start of the monster span or within the monster span or not within the monster span. 

$4520$ sentences of data were obtained by combining all the monster lore data. These sentences were divided into training ($2/3$), development ($1/6$), and test ($1/6$) sets without separating sentences from the same lore into different sets. After this step, the sets were separated with 3011, 764 and 745 sentences respectively. In addition to the test set that was obtained by tagging above, the test set that was obtained during setup 2 is manually verified and made as a gold standard test data for comparing the performance of different models.

Also, it was observed that the \textit{Trankit} tokenizer does not separate certain domain-specific words into tokens. For example, words such as \textit{wyrmling dragon}, \textit{a larder} are not tokenized into two tokens. As FLAIR does not support two tokens with space in \verb|BIO| format, the specific words are separated into 2 lines using the spaces.

\subsection{Training and Evaluation}
The evaluation is done for each setup with different models and the results of the model training for both setups were compared with \textit{Trankit} NER system that is not custom-trained with the lore data (zero-shot Trankit model). Zero-shot Trankit model is used as the baseline model to evaluate the effect of fine-tuning for fantasy domain data. As the zero-shot Trankit model does not have the monster category, it is expected that the monster names are identified as PER category entities. For example, it is taken that the zero-shot model has performed correct recognition if the model recognises \verb|B-MONS| as \verb|B-PER|. Model training and evaluation were done on the datasets that were created using setup 1 and setup 2. Model training and evaluation was done with 1 model configuration in \textit{Trankit} and 2 model configurations in \textit{FLAIR}.

For each experiment with \textit{Trankit}, a custom \verb|Trankit pipeline| was configured for custom-NER for English with train and development set \verb|IOB| tagged files as the input. For this pipeline, pre-trained xlm-roberta-base which was available in \textit{Trankit} is used as the model that is trained with the customized pipeline. Batch size and the number of epochs were the tuned hyperparameters. It was identified that batch size 32 had the better performance for \textit{FLAIR} models and 16 was the best for \textit{Trankit}. Also number of epochs was 25 as identified by hyperparameter tuning. Then, the training was done for 25 epochs, saving the model with the best validation F1 score on the development set. Then, the obtained models were tested with the tagged test set and the gold standard test set and the F1 scores were obtained. 

There were two model configurations that were experimented with FLAIR. Configuration 1 uses the \verb|stacked embeddings| as proposed by~\citet{akbik2018contextual} with BiLSTM sequence tagger. Configuration 2 uses the same \verb|stacked embeddings| with BiLSTM-CRF as the sequence tagger.

After the above steps, the best model was taken and each monster lore was fed into the pipeline created for the best model. Then, the association maps were obtained according to the monsters tagged in each monster's lore. Further, another comparison was done based on the association map obtained by the text lookup against the association map obtained by tagging with the model. In addition to this, a directed graph was created from each association map to visualize the associations between all the monsters.

\section{Results}
\label{sec:results}

Table I shows the results obtained on the test set for the models with data obtained from setup 1 and setup 2.  

\begin{table}[htbp]
 \centering
\caption{Models and their performance with text lookup test data}
\label{tab1}
\begin{tabular}{|l|l|c|c|c|}
\hline
\textbf{Model} & \textbf{\textit{Training}} & \textbf{\textit{Precision}}& \textbf{\textit{Recall}}& \textbf{\textit{F1}} \\
\hline
\multirow{2}{*}{FLAIR Config 1} & Setup 1 & 62.11 & 72.46 & 66.89 \\
\hhline{~----}
&  Setup 2 & 91.91 & 92.46 & 92.19 \\
\hline
\multirow{2}{*}{FLAIR Config 2} & Setup 1 & 64.91 & 80.43 & 71.84 \\
\hhline{~----}
& Setup 2 & 94.11 & \textbf{92.87} & 92.97 \\
\hline
\multirow{3}{*}{Trankit} & Zero-shot  & 00.00 & 00.00 & 00.00 \\
\hhline{~----}
& Setup 1 & 66.42 & 65.94 & 66.18 \\
\hhline{~----}
& Setup 2 & \textbf{96.67} & 92.26 & \textbf{94.42} \\
\hline

\hline

\multicolumn{4}{l}{}
\end{tabular}
\end{table}

\begin{table}[htbp]
 \centering
\caption{Models and their performance evaluated for gold standard test set}
\label{tab2}
\begin{tabular}{|l|l|c|c|c|}
\hline
\textbf{Model} & \textbf{\textit{Training}} & \textbf{\textit{Precision}}& \textbf{\textit{Recall}}& \textbf{\textit{F1}} \\
\hline
\multirow{2}{*}{FLAIR Config 1} & Setup 1 & 82.97 & 28.23 & 42.12 \\
\hhline{~----}
&  Setup 2 & 85.58 & 89.17 & 87.34 \\
\hline
\multirow{2}{*}{FLAIR Config 2} & Setup 1 & 86.10 & 28.58 & 42.87 \\
\hhline{~----}
& Setup 2 & 85.47 & \textbf{89.59} & 87.43 \\
\hline
\multirow{3}{*}{Trankit} & Zero-shot  & 00.00 & 00.00 & 00.00 \\
\hhline{~----}
& Setup 1 & 82.58 & 26.78 & 40.44 \\
\hhline{~----}
& Setup 2 & \textbf{86.44} & 89.33 & \textbf{87.86} \\
\hline

\hline

\multicolumn{4}{l}{}
\end{tabular}
\end{table}

It was observed that zero-shot \textit{Trankit} was not able to identify any of the monster name tokens as \verb|PER|. Based on the results of the setup 1 and setup 2 experiments, it was found that setup 2 showed better performance than setup 1 as the monster names from the lore only might not have been enough for training. This is the reason for the improvement of performance when using \textit{FRW-I} data.

When comparing the models both the \textit{FLAIR} models, it was observed that the model with LSTM-CRF tagger has outperformed the model with the LSTM tagger. Further, comparing both the \textit{FLAIR} models with the \textit{Trankit} model, it was observed that both the \textit{FLAIR} models have performed better in terms of recall and F1 than the \textit{Trankit} model. Also it was observed that the \textit{Trankit} performs better with Setup 2 in terms of precision and F1 while recall was almost competitive. 

\begin{figure*}[!htbp]
\centerline{\includegraphics[width=\textwidth]{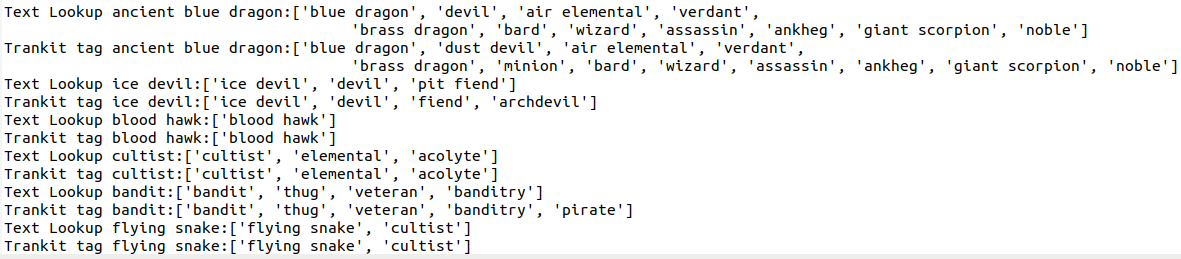}}
\caption{A part of the association map(list) created to compare text lookup and \textit{Trankit} outputs.}
\label{fig}
\end{figure*}

\begin{figure*}[!htbp]
\centerline{\includegraphics[width=\textwidth]{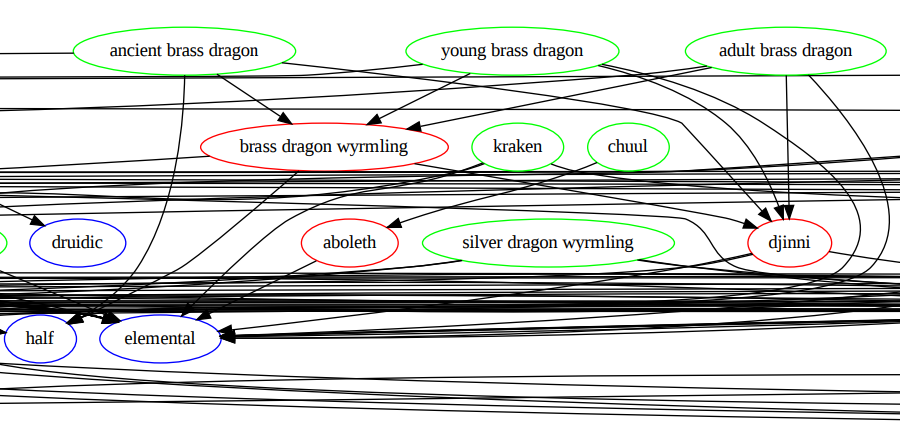}}
\caption{A part of the directed graph obtained from the tagging by \textit{Trankit}}
\label{fig}
\end{figure*}

As a way of comparing the outputs obtained from text lookup and trankit which is the model that performed the best, associations between the monsters were listed for both setups. These association maps were compared to identify the difference while tagging with \textit{trankit}. In the association maps, the following were observed. 
\begin{itemize}
    \item In the Trankit tagged output for the lore data for setup 2, some additional monsters which were in the lore data but, not in the list that was used to tag monsters, were tagged. Also, some monsters which are top level categories of some monsters in the monster list are also identified. \textit{Archdevil} and \textit{dust devil} were identified as additional monsters identified by trankit. Also centipede and pirate are the monster categories that were tagged as monsters.
    \item In trankit output of setup 2, it was found that there are some additional entities that were tagged. These were found not to be monster names but, they represent monsters according to the context that they appear in the text. Fanatic (used to represent cult fanatic), flying dragon (used to represent a dragon) and worm (refers to purple worm) were identified. 
    \item In setup 2 output, \textit{djinn} which is a plural of \textit{djinni} is identified even when not tagged in text lookup.
    \item In setup 2, certain non-monster entities are identified as monster entities. \textit{Ice shard}, \textit{nomad} , \textit{geyser} can be taken as examples. 
    \item Also in setup 2, it was observed there were only a very small number of non-members were identified as monster entities.
    \item In setup 1, there were similar observations made but there were many misses observed where in several mentions of the same entity, some mentions were missed.     
    
\end{itemize}

\section{Conclusion}
\label{sec:conc}

Here, \textit{Trankit} is finetuned using fantasy domain data and achieves an F1 score of 87.86\% with the monster lore data from DnD using fine-tuned \textit{Trankit} as an end-to-end pipeline. Also, it can be also concluded that fine-tuning for fantasy domain data is absolutely necessary as the performance metrics showed 0.00\% with zero shot \textit{Trankit}. Also, it was observed that fine-tuned \textit{Trankit} also identifies entities using context and it can be seen from the identification of the untagged monsters. \textit{FLAIR} models also performed with comparable performance, especially with less data. From the F1 score and the analysis of observations, it can be seen that the models are fine-tuned with considerably better performance for the monster entity recognition.

\bibliography{refernces}
\bibliographystyle{IEEEtranN}

\end{document}